\documentclass[11pt, a4paper]{article}

\usepackage[top=2.5cm, bottom=2.5cm, left=2.5cm, right=2.5cm]{geometry}

\usepackage{amsmath}
\usepackage{amsfonts}
\usepackage{amssymb}
\usepackage{graphicx}
\usepackage{float}
\usepackage{booktabs}
\usepackage{enumitem}
\usepackage[utf8]{inputenc} 
\usepackage{tikz}
\usetikzlibrary{shapes, arrows.meta, positioning, calc, shadows, fit, backgrounds}
\usepackage{xurl}

\usepackage[T1]{fontenc}
\usepackage[utf8]{inputenc}
\usepackage{fancyvrb}
\usepackage[most]{tcolorbox}
\usepackage{amsfonts} 
\usepackage{fvextra} 
\usepackage[most]{tcolorbox}
\tcbuselibrary{breakable, skins}

\usepackage{tabularx}
\newcolumntype{C}{>{\centering\arraybackslash}X}

\newtcolorbox{promptbox}[1]{
    breakable,
    enhanced,
    colback=gray!5,
    colframe=gray!30,
    title={#1},
    fonttitle=\bfseries\sffamily,
    coltitle=black,
    attach boxed title to top left={yshift=-2mm, xshift=2mm},
    boxed title style={colback=gray!20, sharp corners},
    sharp corners,
    fontupper=\small\ttfamily,
}

\usepackage{mathpazo} 

\usepackage{authblk}

\usepackage[square,numbers]{natbib}

\usepackage[colorlinks=true, linkcolor=blue, citecolor=blue, urlcolor=blue]{hyperref}

\usepackage{siunitx}

\title{\textbf{Can LLMs Do Rocket Science? Exploring the Limits of Complex Reasoning with GTOC 12}}

\author[1]{Iñaki del Campo}
\author[1]{Pablo Cuervo}
\author[1]{Victor Rodriguez-Fernandez}
\author[2]{Roberto Armellin}
\author[2]{Jack Yarndley}

\affil[1]{Universidad Politécnica de Madrid (UPM), Spain \protect\\ \texttt{\{i.delcampo, victor.rfernandez\}@upm.es}}
\affil[2]{University of Auckland (UoA), New Zealand \protect\\ \texttt{roberto.armellin@auckland.ac.nz}}

\date{\today} 

\begin{document}

\maketitle

\begin{abstract}
Large Language Models (LLMs) have demonstrated remarkable proficiency in code generation and general reasoning, yet their capacity for autonomous multi-stage planning in high-dimensional, physically constrained environments remains an open research question. This study investigates the limits of current AI agents by evaluating them against the 12th Global Trajectory Optimization Competition (GTOC 12), a complex astrodynamics challenge requiring the design of a large-scale asteroid mining campaign. We adapt the MLE-Bench framework to the domain of orbital mechanics and deploy an AIDE-based agent architecture to autonomously generate and refine mission solutions. To assess performance beyond binary validity, we employ an "LLM-as-a-Judge" methodology, utilizing a rubric developed by domain experts to evaluate strategic viability across five structural categories. A comparative analysis of models, ranging from GPT-4-Turbo to reasoning-enhanced architectures like Gemini 2.5 Pro and o3, reveals a significant trend: the average strategic viability score has nearly doubled in the last two years (rising from 9.3 to 17.2 out of 26). However, we identify a critical capability gap between strategy and execution. While advanced models demonstrate sophisticated conceptual understanding—correctly framing objective functions and mission architectures—they consistently fail at implementation due to physical unit inconsistencies, boundary condition errors, and inefficient debugging loops. We conclude that, while current LLMs often demonstrate sufficient knowledge and intelligence to tackle space science tasks, they remain limited by an implementation barrier, functioning as powerful domain facilitators rather than fully autonomous engineers.
\end{abstract}

\section{Introduction}
\noindent 
Large Language Models (LLMs) have revolutionized AI, yet their capacity for truly complex reasoning and multi-stage planning in severely constrained, real-world problems remains a frontier. Current benchmarks often don't capture this depth, and the rapid advancement of LLMs has quickly outpaced many of the benchmarks we use to evaluate them. In this study, we propose to evaluate LLMs against a recognized optimization challenge: the Global Trajectory Optimization Competition (GTOC). We will focus on GTOC 12: Asteroid mining\cite{zhang2025sustainable}.
This challenge involves designing asteroid mining missions, a problem demanding intricate trajectory planning, strategy, resource optimization, and even novel ideas for successful resolution. Despite interest in LLMs, there's a gap in their systematic evaluation within domains of this magnitude.
Our methodology adapts GTOC 12 to an interactive environment based on OpenAI's mle-bench, a benchmark for AI agents at machine learning tasks, acknowledging that GTOC is a numerical and strategic optimization problem, not traditional machine learning. This will require defining a clear interface and validation system. Subsequently, we will develop an LLM agent based on the AIDE (Agent for Iterative Data-science Exploration) architecture \cite{jiang2025aide}. AIDE is an agent architecture that automates machine learning engineering by systematically drafting, debugging, and refining solutions through a tree greedy search aproach. The agent will be developed iteratively through prompt engineering, modifications to its operational flow, and the provision of external tools. On the one hand, performance will be measured by its ability to generate valid proposals, and scoring will be based on the formula proposed by the competition. On the other hand, a qualitative assessment of the proposed solutions will be carried out, for which LLM judges will apply a correction template designed by expert competitors in that edition. 
This study seeks to answer crucial questions: 
\begin{itemize}
    \item What is the baseline performance of current LLMs on such complex tasks?
    \item What difficulties cause LLMs to fail in this challenge?
    \item Do they typically fail due to poor strategy, or are they capable of properly framing the problem, with implementation being their limitation?
    \item Can they iteratively improve proposed solutions for such complex problems?
    \item How do prompting strategies, agent design, and external tools impact their performance?
\end{itemize}
Ultimately, we aim to determine how far current LLM technology can go in solving problems that traditionally demand deep human expertise and considerable optimization computation.

\section{Background}
This section establishes the foundational context for the study, bridging the gap between general-purpose AI benchmarks and domain-specific engineering challenges. We first review MLE-Bench and the AIDE framework to characterize the current state-of-the-art in autonomous machine learning engineering. Subsequently, we introduce the GTOC 12 competition, defining the high-dimensional optimization environment used to evaluate the transferability of these agents to the field of astrodynamics.\\

\subsection{GTOC and GTOC 12: The Cup of Rocket Science}
GTOC (Global Trajectory Optimisation Competition) is the premier international competition for interplanetary trajectory design. Initiated by the European Space Agency (ESA) in 2005, it challenges the world's best aerospace engineers and mathematicians to solve ``nearly impossible'' orbital mechanics problems within a roughly four-week timeframe. A unique tradition of the competition is that the winning team of each edition is tasked with designing the problem for the following one. For this research, we will focus on the twelfth edition of GTOC 12, Sustainable Asteroid Mining:

\paragraph{GTOC 12: Sustainable Asteroid Mining \cite{zhang2025sustainable}}
The 12th edition (2023) was the latest edition up to the start of this research. Organized by Tsinghua University, the problem focused on a futuristic scenario where sustainable asteroid mining is put into action.

\begin{figure}[H]
    \centering
    \includegraphics[width=0.7\linewidth, height=0.5\textheight, keepaspectratio]{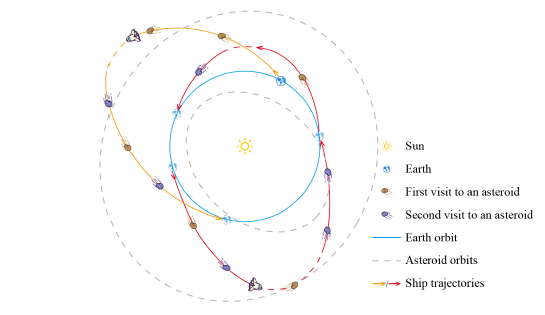}
    \caption{Schematic of asteroid mining by multiple Mining Ships\cite{zhang2025sustainable}}
    \label{fig:gtoc_12}
\end{figure}

\begin{itemize}
    \item \textbf{The Problem:} Teams were tasked with designing trajectories for a fleet of ``Mining Ships'' launching from Earth between the years 2035 and 2050. The objective was to transport as many asteroid resources back to Earth as possible within this 15-year window. The ships must rendezvous with targets selected from a candidate list of 60000 asteroids to extract and refine valuable minerals.
    \item \textbf{Complexity:} The challenge is a massive combinatorial optimization puzzle. It requires selecting which asteroids to visit (from thousands), in what sequence, and with which ship, all while strictly adhering to physical constraints like low-thrust electric propulsion limits and flight duration. It serves as a really high-complexity benchmark for global optimization algorithms.
\end{itemize}

\subsubsection{GTOC 12 score function}

The primary objective of the GTOC 12 ``Sustainable Asteroid Mining'' challenge is to maximize the total resource value collected from asteroids and transported back to Earth within a 15-year mission window (2035--2050). The global merit function $J$ is defined as the weighted sum of the mass returned from each mined asteroid:

\begin{equation}
    J = \sum_{i=1}^{60,000} B_i M_i
\end{equation}

where $M_i$ represents the collected mass (in kilograms) from the $i$-th asteroid, and $B_i$ is a bonus coefficient associated with that target.

In the official competition, the problem includes a game-theoretic element where the bonus coefficient $B_i$ is dynamic. It decays in real-time based on the cumulative mass mined by all competing teams on the global leaderboard, governed by the following decay function:

\begin{equation}
    B_{i} = \frac{1+2(1+\beta\sum M_{i})^{\gamma}}{3}
\end{equation}

where $\beta=0.05~kg^{-1}$ and $\gamma=-0.1$. However, for the purpose of this paper, we evaluate the solution locally rather than in the live competitive environment. Consequently, we strictly eliminate this game-theoretic element. We assume a static environment where the bonus coefficient remains constant (i.e., $B_i = B_0 = 1$) to isolate the trajectory optimization performance from external adversarial factors.

The collected mass per asteroid, $M_i$, remains strictly constrained by the duration of the mining operation. Successful extraction requires two successive rendezvous events (deployment and retrieval), where the yield is proportional to the stay duration $\Delta t = t_{2i}-t_{1i}$:

\begin{equation}
    M_{i} \le k(t_{2i}-t_{1i})
\end{equation}

Here, the minimum stay time is 1 year, and the mining rate is fixed at $k=10.0~kg/yr$.

Finally, the scale of the operation is constrained by efficiency. The maximum allowable number of Mining Ships, $N$, is dynamically capped based on the solution's average collected mass per ship ($\overline{M}$):

\begin{equation}
    N \le \min[100, 2\exp(\rho \overline{M})]
\end{equation}

where $\rho = 0.004~kg^{-1}$. This forces the optimization agent to prioritize high-yield targets, as low-yield missions negatively impact the allowable fleet size.\\

\subsubsection{Trajectory and State Constraints}
All mission events, including launches, asteroid rendezvous, planetary flybys, and resource returns, must occur within the specified epoch range of January 1, 2035 (64328 MJD) to January 1, 2050 (69807 MJD).\\\\
The heliocentric trajectories are constrained by the following boundary conditions:\\\\
\textbf{Launch and Return:} Spacecraft must match Earth's position. The hyperbolic excess velocity relative to Earth, $v_{\infty}$, is constrained at both launch and unloading events:
If a ship performs an Earth flyby with $v_{\infty} \le 6.0 ~ km/s$, all collected resources are instantaneously unloaded.
    \begin{equation}
        v_{\infty} \le 6.0 ~ km/s
    \end{equation}
\\
\textbf{Rendezvous:} The spacecraft position and velocity must strictly match that of the target asteroid (relative velocity equal to zero).\\\\
\textbf{Solar Proximity:} The distance from the Sun $r(t)$ must remain within safe thermal limits throughout the mission:
    \begin{equation}
        r(t) \ge 0.3 ~ AU
    \end{equation}
    where $1 ~ AU = 1.49597870691 \times 10^8 ~ km$.

\subsubsection{Spacecraft and Propulsion Model}
Each Mining Ship acts as a variable-mass system propelled by low-thrust electric propulsion. The motion is governed by the standard equations of motion perturbed by the thrust vector $\mathbf{T}$:

\begin{equation}
    \dot{\mathbf{v}} = -\frac{\mu}{r^3}\mathbf{r} + \frac{T}{m}\boldsymbol{\alpha}
\end{equation}

where $\boldsymbol{\alpha}$ is the thrust direction unit vector. The propulsion system is characterized by a specific impulse $I_{sp} = 4000 ~ s$ and a maximum thrust $T_{max} = 0.6 ~ N$. The mass depletion rate is defined by:

\begin{equation}
    \dot{m} = -\frac{T}{I_{sp} g_0}
\end{equation}

where $g_0 = 9.80665 ~ m/s^2$. The initial wet mass of each ship, $m_0$, including the dry mass ($m_d=500~kg$), propellant, and miner payload ($m_s=40~kg$ per miner), must not exceed $3000 ~ kg$.

\subsubsection{Gravity Assist Model}
To improve reachability, ships may perform gravity assists (GA) at Venus, Earth, and Mars. We utilize a powered flyby model where the interaction is treated as an instantaneous change in the velocity vector's direction, while the magnitude of the hyperbolic excess velocity $v_{\infty}$ remains constant.\\

The maximum deflection angle $\delta$ is constrained by the minimum safe pericenter radius $r_{p,min}$ of the specific planet:

\begin{equation}
    \sin\left(\frac{\delta}{2}\right) = \frac{\mu_p / r_{p,min}}{v_{\infty}^2 + \mu_p / r_{p,min}}
\end{equation}

where $\mu_p$ is the planet's gravitational parameter. The minimum pericenter radii are $6351 ~ km$ for Venus, $6678 ~ km$ for Earth, and $3689 ~ km$ for Mars.

\subsection{MLE-Bench: Evaluating AI Agents on Machine Learning Engineering}
MLE-Bench\cite{chan2024mle} is a benchmark introduced by OpenAI to rigorously evaluate the capabilities of AI agents in performing end-to-end Machine Learning Engineering (MLE) tasks. Unlike traditional coding benchmarks (e.g., HumanEval\cite{chen2021evaluating} or SWE-bench\cite{jimenez2023swe}) that focus on isolated functions or general software maintenance, MLE-Bench targets the open-ended, scientific workflow of a data scientist. The key characteristics and methodology of MLE-Bench are detailed below:

\begin{itemize}
    \item \textbf{Task Composition:} The benchmark consists of 75 competitions curated from Kaggle \cite{kaggle2024}, a premier platform for data science challenges. Spanning diverse domains such as computer vision, natural language processing, and tabular data analysis, these tasks require agents to manage the entire machine learning pipeline, including data preparation, architecture design, hyperparameter tuning, and the generation of valid submission files. Tasks are categorized by complexity: Low (solvable by a human in <2 hours), Medium (2–10 hours), and High (>10 hours)
    \item \textbf{Agent Environment:} Agents are provided with a competition description, a dataset, and a local validation server. They must autonomously perform data preparation, model training, and experimentation to produce a standardized CSV submission.
    \item \textbf{Evaluation:} Since every challenge has their own evaluation metric, the performance is contextualized using real-world human leaderboards. The framework applies Kaggle’s specific medal thresholding logic (Bronze, Silver, and Gold) to agent submissions, enabling a direct comparison between AI autonomy and historical human-level achievement.
\end{itemize}

\subsection{AIDE: AI-Driven Exploration}
AIDE (Agent for Iterative Data-science Exploration) is the scaffolding framework that achieved state-of-the-art results on the MLE-Bench, particularly when powered by reasoning models like OpenAI's o1-preview\cite{openai2024o1}. It represents a paradigm shift from simple ``Chat-and-Execute'' agents to a structured, search-based approach.

\begin{itemize}
    \item \textbf{Architecture (Tree Search):} AIDE treats the MLE process as a search problem within a space of solution candidates. Instead of a single linear attempt, it constructs a solution tree. Each node in the tree represents a distinct version of the solution (e.g., a specific model architecture or preprocessing strategy).
    \item \textbf{Iterative Refinement:} The agent utilizes a ``Best-First Search'' strategy to navigate this tree. It drafts an initial solution, executes it, and analyzes the results (error logs or validation metrics). Based on this feedback, it ``branches'' out by proposing specific improvements or bug fixes.
    \item \textbf{Relation to Bench:} On MLE-Bench, AIDE significantly outperformed other frameworks. Its success demonstrates that scaffolding, the structural logic governing how a model plans, backtracks, and learns from failure, is as critical as the underlying LLM's intelligence for complex engineering tasks.
\end{itemize}

\section{Methodology: Adapting MLE-Bench for Astrodynamics}

To evaluate the capabilities of Large Language Models in complex trajectory optimization, we required a robust testing framework that simulates an autonomous engineering workflow. We selected MLE-Bench as our foundation due to its sandboxed proved approach to data science competitions. However, because MLE-Bench is inherently designed for standard Machine Learning tasks, where success is measured by prediction accuracy on a test set, significant adaptations were necessary to accommodate the GTOC 12 challenge. This section details the technical modifications made to both the environment and the AIDE agent architecture to bridge the gap between classification or regression tasks and orbital mechanics.

\subsection{Environment Adaptation: From Kaggle to Trajectory Optimization}

The first phase of our methodology involved transforming the standard MLE-Bench task structure into a GTOC-compatible environment. We created a custom challenge that mirrors the structure of a Kaggle competition but fundamentally alters the input and output mechanisms. The original problem statement and submission guidelines, originally in PDF format, were digitized and simplified into Markdown. This ensures that the agent receives clear, token-efficient text instructions without the noise of visual formatting. Furthermore, we rewrote the environment's inherent system instructions to remove all references to training sets, test splits, or model inference, orienting the agent instead towards numerical optimization and constraint satisfaction.\\

A critical modification was the integration of the domain-specific validation system. In a standard MLE-Bench task, the agent produces a CSV of predictions; in our adaptation, the agent must generate a strictly formatted solution file containing trajectory data. We integrated the official GTOC 12 trajectory validator directly into the evaluation loop as a subprocess. This allows the system to provide immediate, ground-truth feedback on physical validity and scoring, replacing standard ML metrics like accuracy or F1-score. Finally, we rebuilt the underlying Docker environment to support this workflow. We replaced standard deep learning dependencies with a suite of scientific computing and astrodynamics libraries, and introduced a mechanism that allow us to initialize the agent to search from previous draft seed solution rather than starting from scratch every iteration.

\subsection{Agent Refinement: Generalizing AIDE for Physics-Based Reasoning}

The second phase focused on modifying the AIDE agent, which, in its default state, exhibits a strong inductive bias towards hyperparameter tuning and model training. To repurpose it for physics simulations, we first extracted all hardcoded prompts embedded within the agent's codebase into an external configuration file (prompts.txt). This modularity allowed us to overwrite the agent's internal monologue, effectively transforming it from a Data Scientist into a Guidance, Navigation, and Control Engineer, without altering the core search logic. We also adapted the feedback functions to parse the output of the GTOC validator, enabling the agent to interpret error logs related to orbital mechanics rather than loss convergence.\\

On the software engineering side, we addressed several limitations to improve the agent's reasoning capabilities. We implemented a fix for a critical bug in the MLE-Bench integration where terminal output was not correctly passed to subsequent search nodes, a failure that previously caused hallucinations regarding code execution status. We also added an automated handler to correct common Python entry-point errors (specifically the \_\_main\_\_ block execution), ensuring that the agent's scripts were executable in a headless environment. To facilitate iterative experimentation, we added a configurable feature to pre-load specific nodes from the search tree, allowing us to resume experiments or force the agent to refine a specific human-generated strategy.\\

Finally, to mitigate the most frequent failure modes observed during preliminary testing, we injected advices into the system prompt. The agent is explicitly instructed to prioritize computational efficiency given the time constraints, urging it to minimize search spaces rather than attempting brute-force calculations. We also enforce strict protocols for debugging—requiring the agent to fix one error at a time—and include warnings about specific technical pitfalls, such as the consistency of physical units across libraries and the handling of trailing newlines in solution files, which frequently caused the validator to crash.

\section{Evaluation Framework}

\subsection{Plan Evaluation Rubric}

As of today, given the stochastic nature of Large Language Models (LLMs) and the complexity of the GTOC 12 problem, relying on an agent to immediately generate a fully valid, high-scoring solution file is generally infeasible. The gap between a plausible-sounding strategy and a mathematically valid trajectory is significant. In all the conducted attempts, the agent fails to produce a solution that passes the rigorous bounds of the validator or scores any points at all. \\
\\
For that reason, we propose an evaluation of the general strategy provided by the LLM. To this end, a team of experts from the University of Auckland, who have been participating and researching this challenge \cite{armellin2025gtoc12}, have established a correction template that decomposes the proposed plan into five distinct structural categories. This helps us to understand the limitations of current models by observing errors in their strategy and allows us to evaluate the degree of intelligence of different LLMs in this task. \\
\\
This evaluation is performed by a second LLM, gemini 2.5 pro, acting as a judge that reviews the draft against a specific rubric of 26 criteria. Each criterion is evaluated as either \textit{True} or \textit{False} based on the content of the plan. A ``True'' rating awards one point, resulting in a maximum possible viability score of 26. Five categories of the evaluation checklist are defined as follows:

\begin{table}[H]
    \centering
    \caption{Evaluation Criteria for Model-Generated Solutions}
    \label{tab:eval_criteria}
    \small 
    \renewcommand{\arraystretch}{1.3} 
    \begin{tabularx}{\linewidth}{@{}lX@{}} 
        \toprule
        \textbf{Category} & \textbf{Evaluation Criteria} \\
        \midrule
        
        \textbf{1. Objective Function} & 
        \begin{minipage}[t]{\linewidth}
            \begin{itemize}[leftmargin=*, nosep, after=\vspace{\baselineskip}]
                \item Maximize mined mass via immediate deployment strategies.
                \item Prioritize late collection to maximize mining duration.
                \item Correlate mission count limits with total mined mass.
                \item Quantify impact of transfer time savings on mass yield.
            \end{itemize}
        \end{minipage} \\
        
        \textbf{2. Asteroid Filtering} & 
        \begin{minipage}[t]{\linewidth}
            \begin{itemize}[leftmargin=*, nosep, after=\vspace{\baselineskip}]
                \item Filter inaccessible targets (costly Earth$\to$Ast or Ast$\to$Earth).
                \item Robust ranking methodology based on transfer costs ($\Delta V$).
                \item Cluster targets reachable early (deployment phase).
                \item Cluster targets viable for late return (collection phase).
                \item Strategy for asteroid-to-asteroid ($A \to A$) clustering.
            \end{itemize}
        \end{minipage} \\

        \textbf{3. Low-Thrust Est.} & 
        \begin{minipage}[t]{\linewidth}
            \begin{itemize}[leftmargin=*, nosep, after=\vspace{\baselineskip}]
                \item Distinguish high-fidelity limitations for preliminary design.
                \item Fast, approximate feasibility checks (filtering).
                \item Optimization method for low-thrust transfers with required accuracy.
                \item Transcription methodology to competition solution format.
            \end{itemize}
        \end{minipage} \\

        \textbf{4. Architecture} & 
        \begin{minipage}[t]{\linewidth}
            \begin{itemize}[leftmargin=*, nosep, after=\vspace{\baselineskip}]
                \item Minimum feasible structure (Earth--Asteroid--Wait--Earth).
                \item Single mothership usage for centralized deploy/collect.
                \item Adherence to mass-dependent spacecraft constraints.
                \item Phasing strategy (Early Deploy / Late Collect).
                \item Definition of logical time of flight limits and asteroid capacity per ship.
                \item Combinatorial solver for feasible transfer sequences.
                \item Scaling method from coarse estimates to high-fidelity.
            \end{itemize}
        \end{minipage} \\

        \textbf{5. Coupling \& Opt.} & 
        \begin{minipage}[t]{\linewidth}
            \begin{itemize}[leftmargin=*, nosep] 
                \item Handling of dependencies (Deployers vs. Collectors).
                \item Evaluation of Multi-Gravity Assist (MGA) relevance.
                \item Full optimization of independent vs. interdependent missions.
                \item Global selection algorithm to maximize fleet objective function.
            \end{itemize}
        \end{minipage} \\
        \bottomrule
    \end{tabularx}
\end{table}



\subsection{LLM as a Judge Validation}
The deployment of Gemini 2.5 Pro as an autonomous evaluator (LLM-as-a-judge) necessitates a rigorous validation framework to ensure that the observed performance trends reflect genuine strategic capability rather than artifacts of stochastic variance or systemic bias. While automated assessment offers a scalable alternative to human evaluation, the literature identifies two primary risks in this paradigm: (i) intra-rater inconsistency, arising from the probabilistic nature of LLM inference, and (ii) the self-preference bias, where models may disproportionately favor outputs that mirror their own latent distribution or linguistic style.\\
\\
To establish the benchmark’s credibility , we implement two tests. First, we quantify the statistical stability of the judge’s scores through an Intraclass Correlation Coefficient (ICC) analysis across multiple inference cycles. Second, we perform a blind cross-judging experiment using secondary models to detect and measure potential model-specific favoritism. 

\subsubsection{Evaluation Consistency: Intra-rater Reliability}

To validate the reliability of the LLM-as-a-judge framework, we conducted a consistency analysis using Gemini 2.5 Pro as the autonomous evaluator. We selected the first 20 solutions generated by GPT-5 and subjected each to five independent evaluation cycles using default configuration parameters. Reliability was quantified through the Intraclass Correlation Coefficient (ICC), specifically employing a two-way random-effects model for absolute agreement. The analysis yielded a single-rater consistency of $ICC(2,1) = 0.748$ with a $95\%$ confidence interval (CI) of $[0.59, 0.87]$, representing a high-moderate to good level of reliability for individual assessments. Furthermore, the reliability of the averaged scores across the five runs reached an excellent level of $ICC(2,k) = 0.937$ ($95\%$ CI: $[0.88, 0.97]$), with high statistical significance ($p < 10^{-18}$). These results demonstrate that while individual judgments exhibit inherent stochasticity, the aggregate scoring mechanism provides a highly robust and stable metric for benchmarking complex astrodynamics strategies. However, because our tests evaluate a large number of solutions for each model, we consider that the individual evaluation is representative of the capabilities of the models and the general trend we are trying to measure.\\

The results demonstrated that, by utilizing a model with sufficient reasoning capabilities, the application of the template is consistent, and score variations for the same solution evaluation are marginal. Based on these findings, Gemini 2.5 Pro was selected as the judge model, given its high inference capacity and the contextual understanding necessary to accurately evaluate complex solutions.\\

\subsection{Self-Preference and Cross-Judging Analysis}

To ensure the validity of the ``LLM-as-a-judge'' framework used in the longitudinal study, we conducted a cross-evaluation experiment to quantify systematic biases. While Large Language Models can exhibit self-preference or varying levels of leniency, a robust evaluation requires the judge's scoring to align with a broader consensus. We selected a small-scale validation sample of 10 solution drafts from each model ($n=40$) and performed a full reciprocal evaluation (160 total assessments). For this analysis, Gemini 2.5 Pro, Gemini 3 Pro Preview, O3-2025-04-16, and GPT-5 were selected due to their advanced reasoning capabilities and their demonstrated proficiency in comprehending and evaluating complex solutions.\\

We define the \textit{Self-Preference Delta} ($\Delta$) for a model $m$ as the difference between the score the model assigns to its own drafts ($S_{m,m}$) and the mean score assigned to those same drafts by the other independent judges ($S_{j,m}$):

\begin{equation}
    \Delta_m = S_{m,m} - \frac{1}{J-1} \sum_{j \neq m} S_{j,m}
\end{equation}

The results of this analysis, presented in Table \ref{tab:bias_analysis}, serve to calibrate the reliability of our scoring system.

\begin{table}[H]
    \centering
    \caption{Cross-Evaluation Matrix and Bias Summary (Scores out of 26)}
    \label{tab:bias_analysis}
    \small
    \begin{tabularx}{\linewidth}{lCCCC}
        \toprule
        \textbf{Judge $\downarrow$ / Author $\rightarrow$} & \textbf{Gemini 2.5  pro} & \textbf{Gemini 3 pro} & \textbf{o3-2025-04-16} & \textbf{GPT-5} \\
        \midrule
        Gemini 2.5 Pro & \textbf{16.00} & 13.67 & 19.63 & 15.88 \\
        Gemini 3 pro & 19.00 & \textbf{14.83} & 20.69 & 19.75 \\
        o3-2025-04-16 & 15.50 & 12.67 & \textbf{13.88} & 13.06 \\
        GPT-5 & 14.00 & 10.83 & 13.19 & \textbf{10.75} \\
        \midrule
        \textbf{Received Mean} & 16.13 & 13.00 & 16.84 & 14.86 \\
        \textbf{Self-Pref. Delta ($\Delta$)} & \textbf{-0.17} & +2.44 & -3.96 & -5.48 \\
        \bottomrule
    \end{tabularx}
\end{table}

This validation step provides three key methodological conclusions:
\begin{itemize}
    \item \textbf{Objectivity of the Evaluation Baseline:} \textit{Gemini 2.5 Pro} exhibits a near-zero delta ($\Delta = -0.17$), demonstrating that its scoring is remarkably aligned with the independent consensus of other high-reasoning models. This justifies its use as the primary evaluator.
    \item \textbf{Inter-Model Scoring Stability:} Despite the limited sample size ($n=10$ per model), the scores for each model remain within a consistent range across different judges. This suggests that the evaluation rubric is interpreted with high semantic stability, regardless of the judge's internal architecture.
    \item \textbf{Calibration of Strategic Assessment:} The small-scale results show that while absolute scores may fluctuate depending on a judge's inherent leniency (e.g., GPT-5 being more stringent), the \textit{relative} performance levels remain stable. This confirms that our methodology is reliable for identifying broad trends in strategic reasoning capabilities, even if implementation-level failures remain a common bottleneck for all models.
\end{itemize}

\section{Experiments}

\subsection{Drafts Generation and Debug Attempts}
This section presents the empirical evaluation of the proposed agentic framework applied to the GTOC 12 challenge. The investigation characterize the operational capabilities and limitations of LLM Agents in this domain, analyzing the performance of the fully autonomous agent, focusing on its ability to generate initial drafts and its efficacy in self-correcting code through iterative debugging. This phase establishes a taxonomy of failure modes, highlighting critical barriers in logical reasoning and physical unit consistency.\\

\subsubsection{Self-Debug Attempts}

This subsection outlines the initial experimental phase conducted within the GTOC-Bench environment using the adapted AIDE agent. The first objective was to generate a corpus of 100 initial solutions to serve as seed material for future experiments. For this task, \textbf{Gemini 2.5 Pro} was selected as the generative model, representing the optimal balance between state-of-the-art performance and cost at the time of the investigation.  At this stage, the complexity of the GTOC 12 challenge and the operational boundaries of Large Language Models within this specific domain were not yet fully characterized. Consequently, a two-phase experimental design was adopted to probe these limits.
\paragraph{Phase 1: Exploratory Debugging}
The first experiment involved subjecting the first 50 generated drafts to a constrained debugging process, limited to a maximum of 5 attempts per draft. The expectation was not to achieve fully valid solutions, but to identify the most immediate barriers to code execution. Results showed a prevalence of superficial syntax and compliance errors. The most frequent issues included Python indentation failures and the hallucination of non-existent methods or arguments in external libraries.

\paragraph{Phase 2: Deep Debugging and Error Analysis}
To gain a more granular understanding of the models' reasoning failures, a second, more intensive experiment was designed. The top 20 initial solutions generated by Gemini 2.5 Pro were selected and subjected to an extended debugging cycle of up to 20 attempts. A manual analysis of the resulting 600+ iterations allowed us to identify and categorize the recurrent patterns of failure:

\begin{itemize}
    \item \textbf{Algorithmic Complexity vs. Resources:} The models frequently failed to account for computational complexity, attempting to implement naive algorithms that exceeded time or memory constraints, rather than simplifying the problem.
    
    \item \textbf{Buggy Debug:} A common behavior observed was instead of isolating and fixing a specific reported error, the agent often introduced multiple unrelated changes in a single iteration. This tendency frequently resulted in regressions, where a fix for one bug introduced several new ones.
    
    \item \textbf{Formatting Overfitting (The Trailing Newline):} A surprisingly persistent error involved the injection of an empty line at the end of the solution file. Models appeared overfitted to appending newline characters (\texttt{\textbackslash n}) incrementally when writing files. This seemingly trivial formatting issue caused the competition validator to crash. Notably, the model exhibited a strong bias towards this pattern and often struggled to revert it despite repeated error messages.
    
    \item \textbf{Unit and Coordinate System Inconsistency:} This was identified as the most critical and pervasive source of error. The models consistently struggled to maintain uniformity across physical units (mixing km/m, days/seconds) and reference frames. A frequent failure mode involved mixing standard units defined in the problem statement with default units from external libraries (e.g., JPL Ephemerides) without performing the necessary conversions.
    
    \item \textbf{Lack of Observability and Blind Heuristics:} When faced with logical errors such as a Lambert solver failing to find a solution—the models rarely opted to instrument the code with logging or print statements to inspect variables. Instead, they tended to enter ``absurd loops'' blindly increasing search spaces or modifying parameters without empirical evidence. For instance, a Lambert failure caused by a unit mismatch (e.g., passing AU instead of km) would often be misinterpreted by the model as a need to widen the launch window, leading to fruitless iterations.
\end{itemize}

\subsection{Model Strategy Analysis}
Regarding the generative models under study, a set including GPT-4-tubo\cite{openai2023gpt4turbo}, GPT-4o\cite{openai2024gpt4o}, o1\cite{openai2024o1}, deepseek-R1\cite{guo2025deepseek}, o3\cite{openai2024o3}, Gemini 2.5 Pro\cite{google2025gemini25pro}, GPT-5\cite{openai2025gpt5} and cloud-sonnet-4.5\cite{anthropic2025claudesonnet45} were selected. All reasoning models are configured with the default reasoning effort parameters. This selection stems from a strategic interest in using the posed problem as a benchmark to delineate the current capabilities and limitations of LLMs when facing tasks of this nature. Furthermore, the generational diversity of the chosen models allows us to observe the evolution of performance and reasoning over time, contrasting previous architectures with state-of-the-art models.\\
\\
Finally, to mitigate the effects of stochastic variance—stemming from both the generator and the evaluator—and to ensure the statistical robustness of the results, the experimental design involves the generation of 100 initial plans for each evaluated model. Although slight fluctuations may occasionally occur, experiments confirm a notable structural consistency in the evaluations. Moreover, to ensure comparative fairness, the same judge model (Gemini 2.5 Pro) was applied invariably to all generated drafts, allowing for a reliable interpretation of the general behavior of the strategies.\\

\begin{figure}[H]
    \centering
    \includegraphics[width=1\linewidth, height=0.9\textheight, keepaspectratio]{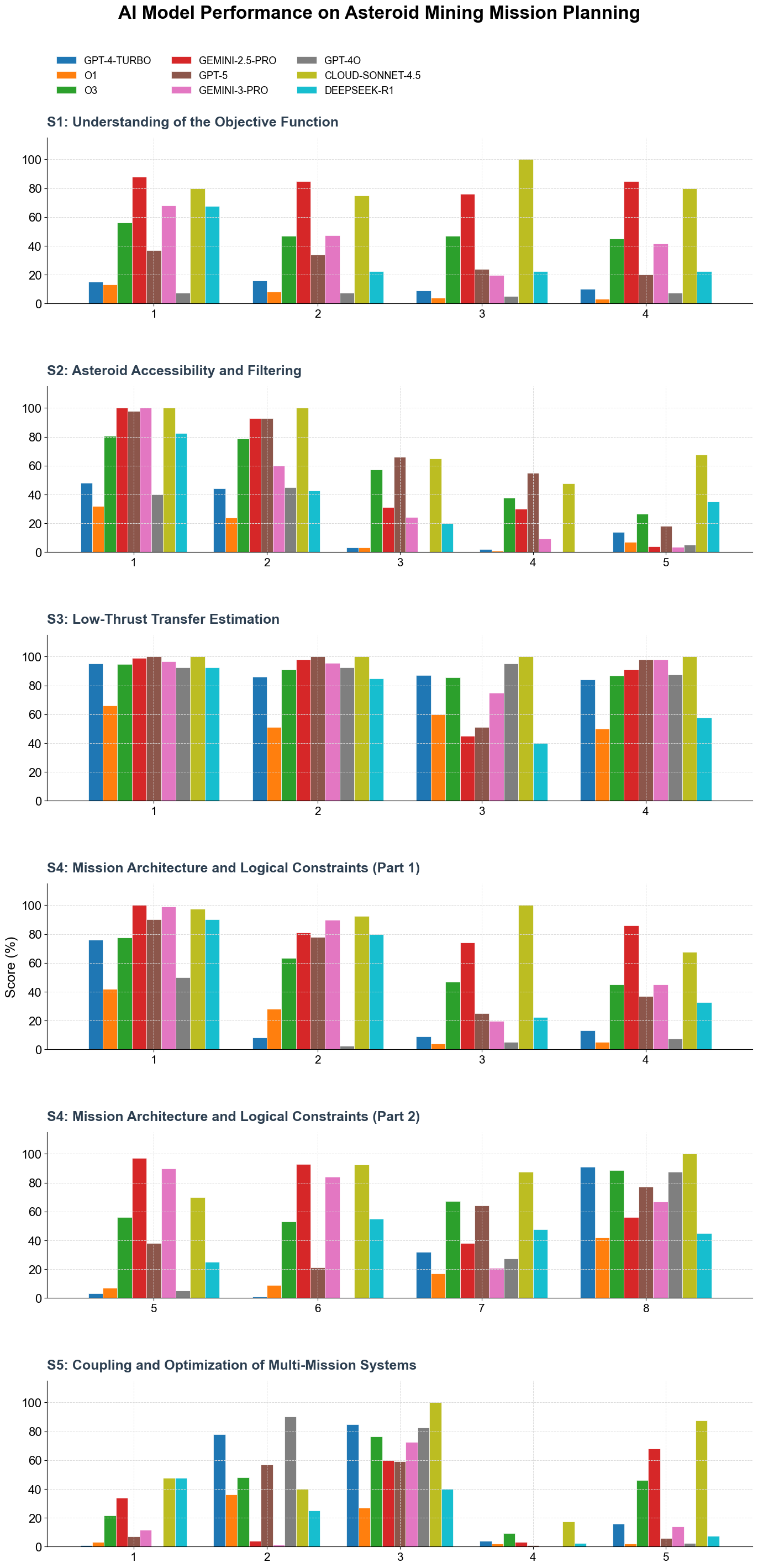}
    \caption{Experiments Results}
    \label{fig:gpt-5}
\end{figure}

\begin{table}[H]
\caption{Comparison of Average Scores by Section across Models}
\label{tab:model_section_comparison}
\begin{tabular}{lcccccc}
\toprule 
Model & Section 1 & Section 2 & Section 3 & Section 4 & Section 5 & Overall Avg \\
\midrule
GPT-4-TURBO & 12.50\% & 22.20\% & 88.00\% & 29.12\% & 36.80\% & 9.30 \\
GPT-4O & 6.88\% & 18.00\% & 91.88\% & 23.12\% & 35.00\% & 8.45 \\
O1 & 7.00\% & 13.40\% & 56.75\% & 19.25\% & 14.00\% & 5.46 \\
DEEPSEEK-R1 & 33.75\% & 36.00\% & 68.75\% & 49.69\% & 24.50\% & 11.10 \\
O3 & 48.72\% & 56.12\% & 89.54\% & 62.24\% & 40.20\% & 15.33 \\
GEMINI-2.5-PRO & 83.50\% & 51.60\% & 83.25\% & 78.12\% & 33.80\% & 17.19 \\
GPT-5 & 28.75\% & 66.00\% & 87.25\% & 53.75\% & 26.00\% & 13.54 \\
CLOUD-SONNET-4.5 & \textbf{83.75\%} & \textbf{76.00\%} & \textbf{100.00\%} & \textbf{88.44\%} & \textbf{58.50\%} & \textbf{21.15} \\
GEMINI-3-PRO & 43.97\% & 39.31\% & 91.09\% & 64.22\% & 19.77\% & 13.49 \\
\bottomrule
\end{tabular}
\end{table}

\textbf{General Constraints and Temporal Evolution: }First, it is necessary to address the exclusion of earlier models from this comparative study. Despite the historical interest in analyzing the trajectory of LLMs, models prior to GPT-4 possess a context window insufficient for processing the problem statement, formatting guide, and environmental constraints. Consequently, these earlier architectures were non-viable for this task.\\
\\
Observing the included models, a clear evolution in score distribution aligns with their release chronology. However, a notable anomaly is the underperformance of \textbf{o1-preview}. Despite being introduced in late 2024 as a revolutionary reasoning model capable of extended thought processes, it failed to translate this capability into superior performance for this specific benchmark. The trend observed is that o1 model often tends to propose an oversimplified strategy, not as a coherent text but as a series of disconnected and numbered code comments. This leads the evaluator, who considers the plan as a student's exam, to grade it with a 0.
Conversely, the overall trend is positive: the mean score of proposed solutions has \textbf{doubled} from the baseline GPT-4-Turbo to the top-performing reasoning model, cloude-sonnet-4.5, highlighting the significant architectural advancements over the last two years.\\

\subsubsection{Section-by-section analysis}

\textbf{Understanding of the objective function: }Disaggregating performance by category reveals distinct patterns. Section 1, which evaluates the model's understanding of key optimization parameters, shows the most dramatic evolution. Among non-reasoning models, scores improved from 12.5 (GPT-4-Turbo) to 28.5 (GPT-5), representing an increase of over 100\%.\\
\\
However, the divergence becomes most apparent when analyzing reasoning models results. The ``deep thinking'' approach proves fundamental for grasping the mechanisms and interdependencies required to optimize the mining mission. Reasoning models consistently outperform their non-reasoning counterparts, ranging from a mean score of 48\% for o3 to an impressive 83\% for cloud-sonnet-4.5 and Gemini 2.5 Pro. This represents a performance leap of approximately 384-664\% compared to the baseline GPT-4-Turbo, and a 168-291\% improvement over the state-of-the-art non-reasoner, GPT-5.\\
 \\
\textbf{Asteroid Accessibility and Filtering:} Section 2 assesses whether the model employs effective pruning strategies like filtering, sorting, and clustering, to manage the search space (e.g., discarding asteroids that exceed fuel/thrust constraints). Contrary to the previous category, reasoning models do not exhibit a distinct advantage here. Again cloud-sonnet-4.5 achieved the highest mean score (76\%), followed by GPT-5 (66\%), by o3 with (56\%) and Gemini 2.5 Pro (51\%). A granular breakdown of this section explains this finding. While all top-tier models correctly identify the need for filtering and sorting (the foundational criteria), they diverge in the more complex task of \textbf{candidate clustering}. Interestingly, reasoning models, particularly Gemini 2.5 Pro, tended to propose simpler but suboptimal solutions, whereas GPT-5 and cloud-sonnet-4.5 demonstrated a capacity to propose sophisticated clustering strategies.\\\\
\textbf{Low-Thrust Transfer Estimation:} This section evaluates the comprehension of feasibility constraints, specifically the necessity of using low-fidelity methods for rapid trajectory assessment before applying high-fidelity optimization. In this domain, even the older GPT-4-turbo and GPT-4o demonstrates a high level of conceptual understanding, achieving a score of 88\% and 91\%. It is important to note, however, that while models possess the theoretical awareness of these requirements, the practical implementation remains a significant hurdle, particularly regarding the strict formatting constraints imposed by the competition validator.\\
\\
\textbf{Mission Architecture and Logical Constraints:} This section evaluates the model's ability to construct an advance mission architecture, moving beyond individual optimization parameters to the structural design of the solution. The criteria defined in Section 4, ranging from minimum feasible structures (Earth-Asteroid-Earth sequences) and spacecraft phasing to the definition of combinatorial solvers, require a high degree of logical consistency and multi-step planning. A priori we hypothesize that reasoning models should exhibit superior performance in this category. Unlike purely semantic tasks, defining a valid mission architecture involves satisfying multiple interdependent constraints (e.g., identify that in GTOC 12, due to the slow mining speed of the devices and transfers, it is more optimal to deploy several mining devices on a trip than to wait next to one on the asteroid before returning.). The results empirically support this hypothesis, revealing a clear capability gap between standard and reasoning architectures.\\
\\
The baseline model demonstrates competence only in the most fundamental aspects of the mission. It achieves high scores in \textit{Minimum feasible structure} (76\%) and \textit{Scalable method to transform coarse estimates to high-fidelity} (91\%). This suggests that GPT-4-turbo conceptually understands what a mission is (it starts and ends at Earth) and how to refine the coarse estimates. However, it struggles significantly with the complex logical constraints required to make the mission operational. In all other subcategories such as phasing, mass-dependent constraints, or time of flight limits, the model fails to perform, scoring in the 5--10\% range. It lacks the depth to connect these isolated concepts into a functional system.\\
\\
The transition to GPT-5 marks a quantitative and qualitative improvement. The average performance across complex subcategories rises to approximately 40\%. Notably, the model excels in two specific structural optimizations: \textit{Simplification via single mothership for deploy/collect} (76\%) and \textit{Algorithm to build feasible transfer sequences} (65\%). This indicates that GPT-5 begins to recognize the combinatorial nature of the problem and the efficiency of centralized architectures, bridging the gap between basic semantic understanding and complex problem structuring.\\
\\
The introduction of reasoning models marks a decisive breaking point in the evaluation. The barrier observed in earlier models, where sophisticated ideas were ignored, is effectively dismantled. These models demonstrate the ability to handle the full spectrum of architectural requirements. o3 demonstrates solid competence, scoring above 50\% across all subcategories, validating its ability to maintain logical consistency. Gemini 2.5 Pro and cloud-sonnet-4.5 achieve state-of-the-art performance, surpassing 75\% and 85\% on average. It is very interesting to note that while one top model may score almost a 100\% in a subcategory, another may struggle in that same subcategory. This is the case with gemini 2.5 pro, which scores 36\% in subcategory S4.7 “Algorithm to build feasible transfer sequences (combinatorial solver),” while claude-sonnet-4.5 doubles its score. This is interesting because it suggests that the future combination of plans developed by different models could significantly help improve solutions.\\
\\
\textbf{Coupling and Optimization of Multi-Mission Systems: } This section addresses the pinnacle of mission complexity: the global optimization and coupling of multi-mission architectures. Moving beyond the design of individual trajectories, the criteria defined in Section 5 evaluate the ability to construct a cohesive fleet strategy. This involves accounting for logistical inter-dependencies (e.g., distinguishing between deployers and collectors), determining the applicability of Multi-Gravity Assists (MGA), and implementing algorithms to select the optimal combination of motherships to maximize the global objective function.\\
\\
Quantitative analysis reveals that this category represents the current upper bound of Large Language Model capabilities in this domain. Overall performance is modest across the board, with no model achieving a dominant result. \textbf{cloud-sonnet-4.5} secures the highest performance with a score of 58.5\%  followed by o3 with approximately 40\%, however non reasoning models achieve decent scores too. It is interesting that a deep inspection of the subcategories reveals a distinct divergence in problem-solving strategies between reasoning and non-reasoning architectures:

\begin{itemize}
    \item \textbf{Non-Reasoning Model:} Models such as GPT-4-Turbo and GPT-5 demonstrate good performance in categories related to orbital mechanics theory, specifically \textit{Determine relevance of Multi-Gravity Assist (MGA)} and \textit{Fully optimize independent missions}.
    \item \textbf{Reasoning Models:} While they do not outperform older models in theoretical concepts, reasoning models (o3, Gemini 2.5 Pro and cloud-sonnet-4.5) begin to show unique competence in the logical structuring of the fleet. They achieve good scores in \textit{Algorithm to select from fully optimized motherships} and \textit{Account for interdependencies}, while non-reasoning models barely exceed 0\%. This suggests that while they may struggle with the physics-based optimization of joint trajectories, they are better equipped to handle the algorithmic logic required to manage system dependencies and selection criteria.
    
\end{itemize}

\subsection{Global LLMs Performance Trend in Rocket Science Domain} \begin{figure}[H] \centering \includegraphics[width=1\linewidth]{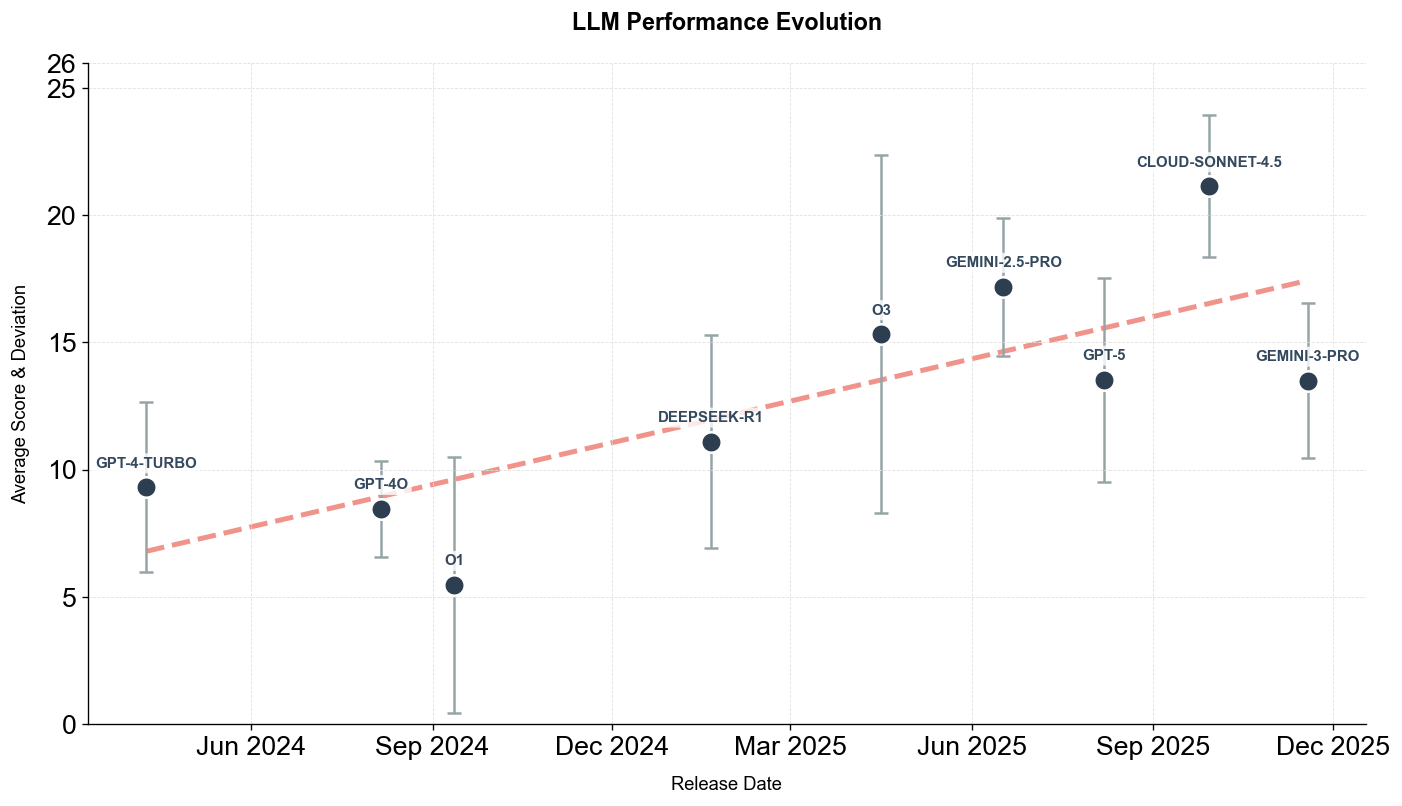} \caption{Performance evolution of LLMs in the GTOC 12 Strategy Evaluation.} \label{fig:trend} \end{figure}

As shown in Figure \ref{fig:trend}, the ability of LLMs to solve complex problems is clearly improving. Based on our evaluation, the average score has nearly doubled in the last two years, rising from 9.3 to 21.15 out of 26. This upward trajectory suggests that the gap between general knowledge and  doing engineering is closing. Although the first models were useful for looking up information, asking questions, or generating ideas, modern reasoning models are beginning to understand the mission as a structured system. If this linear growth continues, the next generation of models may not just propose valid strategies but successfully implement them without the current struggles in formatting and self debugging.

\section{Conclusions}

This study set out to benchmark the capabilities of Large Language Models in the domain of extreme orbital mechanics. We structured our investigation around specific questions regarding performance, failure modes, and the potential for autonomous improvement. Based on our experiments with GTOC 12, we reached the following conclusions:

\paragraph{What is the baseline performance?}
The starting performance of standard LLMs (like GPT-4-Turbo) is low. On their own, they struggle to generate a valid solution for a problem of this magnitude, scoring roughly 9 out of 26 points. However, we observed a massive leap in performance with the introduction of ``reasoning'' models. The ability of a model to “think” before generating text, together with improvements in the models, has allowed the score obtained by the models to double in just over a year and a half. This suggests that raw intelligence is increasing, but the basic level for fully autonomous engineering has not yet been reached.

\paragraph{Strategy vs. Implementation: Why do they fail?}
This was the most critical finding of our research. LLMs typically do not fail due to poor strategy. In fact, advanced models demonstrated a surprisingly strong grasp of the problem. They correctly identified that they needed a mothership, they understood the physics of gravity assists, and they knew how to frame the objective function.
Their limitation is entirely in the implementation. They act like a good architect who can design a perfect building but breaks the bricks when trying to build it. The models consistently failed due to non complex mistakes: syntax errors, mixing up units (km vs. meters), or submission formatting that crashed the validator. They have the right ideas, but for now they lack the precision required for strict fully autonomous software engineering.

\paragraph{Can they iteratively improve their own solutions?}
Currently, the answer is no, but they are close. We have found that without human guidance, LLMs struggle to correct their own mistakes. When an agent encounters an error, it often falls into a “debugging loop”: it tries to correct one mistake, but accidentally introduces two new ones. They often lack the initiative for in-depth “observability,” as they rarely think to add print statements to see what is happening inside the variables. They rely on an interpreter failure to assume that something is wrong. Instead, they guess blindly, which often leads to a dead end.

\paragraph{The Impact of Prompting and External Tools}
The experiments show that the most effective way to improve performance is to learn from the model's failures. Analyzing the most common errors and explicitly warning the model about them in the system prompt proved extremely useful. \\
\\
Ultimately, current LLM technology acts as a powerful facilitator. They have conquered the conceptual barrier of rocket science understanding \textit{what} to do, but they are still halted by the precision barrier of programming code without bugs. They can draft the plan but can not implement it without supervision yet.

\section{Future Work}

The findings of this study open several avenues for research at the intersection of autonomous agents and high-stakes engineering. While we have identified a clear "implementation barrier," future work will focus on bridging the gap between strategic reasoning and execution through three primary pillars:

\begin{itemize} \item \textbf{Architectural Evolution and Self-Correction:} We intend to integrate more sophisticated, evolutionary agent frameworks such as Shinka Evolve \cite{lange2025shinka}. A key priority is the development of "physical-unit-aware" code interpreters and specialized debugging loops that prioritize observability (e.g., automated insertion of telemetry and state-check logging). The goal is to move beyond "blind" trial-and-error toward a systematic refinement process that can handle the strict formatting and physical constraints of GTOC without human intervention. \item \textbf{Expert-in-the-loop and Evaluation Alignment:} A primary limitation of the current study is the reliance on LLM-as-a-judge without a large-scale baseline of expert-reasoned evaluations. Following the evaluation paradigms advocated by \cite{husain2023evals}, future research will incorporate a "Gold Standard" dataset, a collection of mission drafts evaluated and annotated by domain experts with detailed qualitative reasoning. This expert signal will be used to calibrate the LLM judge, identifying where automated scoring diverges from professional intuition and iteratively refining prompt instructions to achieve higher alignment with expert consensus. \item \textbf{Optimization of Expert Solutions:} Beyond generating solutions from scratch, we aim to evaluate the capacity of agents to act as "optimization co-pilots." This involves tasking the LLM with refining valid, high-performing trajectories from previous competition winners. Distinguishing whether AI is limited to "common-sense" orbital mechanics or if it can discover novel, non-intuitive optimizations remains a critical question for the future of autonomous mission design. \end{itemize}

\section*{Code Availability}
{\sloppy
To ensure the reproducibility of our results, all source code, environment configurations, and prompt templates are publicly available. The evaluation environment and the AIDE-based agent architecture are hosted at \href{https://github.com/inaki11/GTOC-Agent-Bench}{github.com/inaki11/GTOC-Agent-Bench} and \href{https://github.com/inaki11/aide-gtoc}{github.com/inaki11/aide-gtoc}, respectively.\par
}
The constituent elements of the dynamic prompts can be located within these repositories as follows: the GTOC 12 mission description and submission requirements reside at \path{/mlebench/competitions/gtoc-upm/description.md}, while the benchmarking system instructions are provided in \path{/environment/instructions.txt}. Furthermore, the specific logic governing agent actions and the "LLM-as-a-Judge" evaluation rubric are detailed in \path{/aide/prompts.txt} and the scoring template.\\
\\
See Appendix \ref{appendix:gtoc_prompt} for initial draft prompt example and Appendix \ref{appendix:judge_prompt} for raw initial correction template.

\section*{Acknowledgements}
This work has been supported by the Spanish Agencia Estatal de Investigacion (AEI) under grant  PID2024-161963OB-C22 (ACTIVATION project). This work has also been supported by the Madrid Government (Comunidad de Madrid-Spain) under the Multiannual Agreement 2023-2026 with Universidad Politécnica de Madrid in the Line A, Emerging PhD researchers.\\
\\
We also want to express our sincere gratitude to Roberto Armellin and his team for their invaluable advice and guidance regarding the 12th Global Trajectory Optimization Competition (GTOC 12) and for their fundamental contribution to the creation of the evaluation template used in this work, which greatly facilitated the analysis presented here.

\bibliography{referencias}

\newpage
\appendix

\section{GTOC 12 Initial Solution Generation Prompt}
\label{appendix:gtoc_prompt}
The following raw Markdown text is an example that was provided as the initial system prompt to the LLM agent. Dynamic variables (e.g., remaining time and steps, agent previous approaches memory) are updated at each iteration of the loop.

\begin{promptbox}{System Prompt: Initial Solution Generation}
\begin{Verbatim}[
breaklines=true, 
    breakanywhere=true, 
    breaksymbolleft={}, 
    breakindent=1.5em, 
    fontsize=\small,     % Replacement for basicstyle
    fontfamily=tt        % Ensures monospaced typewriter font
]

You are an expert spacecraft mission designer + numerical optimizer + software engineer. Your job is to **produce a full, non-naive initial solution generation pipeline for GTOC-12**.

# Task description

You are participating in *GTOC*, a competition where the best aerospace engineers and mathematicians world wide challenge themselves to solve a “nearly-impossible” problem of interplanetary trajectory design, adapted to test the coding ability of agents like yourself.

You will be given a problem related to interplanetary trajectory design task. Your purpose is to produce a valid submission file that must be the result of a trajectory algorithm that controls the spacecraft.

BENCHMARK INSTRUCTIONS
------
- **Data**: The asteroid data files for the competition are available in `data/GTOC12_Asteroids_Data.txt`.
- **Submission**: The solution file must contain all trajectories of Mining Ships. It is a txt file. Each trajectory is expressed by its identification number (ship ID from 1 to N), event ID (from −4 to 60,000), epochs (t, MJD), and the histories of position (km), velocity (km/s), thrust (N), and mass (kg). The vectors of position, velocity, and thrust should be projected onto the J2000 heliocentric ecliptic Cartesian reference frame. 
ADDITIONAL NOTES
------

- **Compute**: You have access to a CPU with the appropriate drivers installed.
- **Total Runtime**: You have a maximum of 10hrs 0mins 0secs hours to attempt this task. You may iteratively develop your solution within this time frame, but bear in mind that program runtime also counts towards this limit.
- **Total Steps**: You have a maximum of 50 steps to submit your solution.
- You will be informed of the remaining time and steps at each step.
- **Use as much of the available time as possible to refine and optimize your submission**: Your goal is to achieve the best possible score, not just to submit your first working solution. Consider strategies like validating your answer using the "Validate Submission" tool at the http://localhost:5000/validate endpoint, manually spot-checking predictions, building custom validation sets and grading functions, and comparing different algorithms. You may keep working until the time limit expires -- as long as your best submission is stored at the designated location at the end of your run, it will be picked up for grading.

COMPETITION INSTRUCTIONS
------

# The Problem of the 12th Global Trajectory Optimisation Competition

## 1. The Scenario

In about 2035, as commercially sustainable asteroid mining may be put into action\*, humanity will plan to mine asteroids and transport their resources back to the Earth. The asteroid mining technology will be greatly developed and integrated into Sustainable Asteroid Miners to automatically extract and refine valuable minerals. A successful mining to an asteroid requires two successive rendezvous to the asteroid by the same or two different Mining Ships propelled by low-thrust electric propulsion. The first rendezvous is to set up a miner and the second is to retrieve the collected mass. Each asteroid can be mined once at most in a submitted solution (i.e., each asteroid can be rendezvoused twice at most). The maximal collected mass mined from an asteroid is proportional to the duration of stay between two moments of rendezvous with this asteroid. The task in GTOC-12 is to transport as many asteroid resources back to the Earth as possible in 15 years.

## 2. The Problem

All mission events must take place between January 1, 2035, 00:00:00 TT (64328 MJD) and January 1, 2050, 00:00:00 TT (69807 MJD) for possible launches, asteroid rendezvous, and planets flybys. Each team can use a limited number of Mining Ships. All Mining Ships must be launched from the Earth, rendezvous with targets selected in the sixty thousand asteroids to set up a miner or to retrieve the collected mass, and fly by the Earth to unload the resources. They can be performed possible gravity assists (GAs) as needed at three terrestrial planets Venus, Earth, and Mars. Each Mining Ship is equipped with an electric propulsion thruster with a specific impulse $I_{sp} = 4000 \text{ s}$ and a maximum thrust magnitude $T_{max} = 0.6 \text{ N}$, while the direction of thrust is not constrained. The Mining Ship's mass changes continuously when the thruster is on, and it must be no smaller than the sum of its dry mass and collected mass throughout the mission. The collected mass unloaded on the Earth is counted in the merit function (see Section 3) and constrains the number of Mining Ships (see Section 4.e)

The candidate asteroids are listed in the file `GTOC12_Asteroids_Data.txt`. They are characterized by identification numbers (asteroid ID from 1 to 60,000), epochs (the mission's initial time 64328 MJD), and corresponding orbital elements (semi-major axis, eccentricity, inclination, longitude of ascending node, argument of perihelion, and mean anomaly).

## 3. Merit Function

The task of the sustainable asteroid mining in this competition is to maximize the total resource value defined by
$$ J = \sum_{i=1}^{60,000} M_i $$
where $M_i$, in kilograms, is the collected mass of the $i$-th asteroid, which is, of course, zero if the collected mass is not transported back to the Earth. 

## 4. Constraints

The scenario of the sustainable asteroid mining mission is set with the following constraints.
- All the mission events (including the launches, thrust on/off, rendezvous, planetary flybys, etc.) must take place between 64328 MJD and 69807 MJD.
- The position of each Mining Ship at the moment of launch, rendezvous, and planetary flybys must be equal to the position of the Earth, asteroid, and planet at the same time, respectively. The hyperbolic excess velocity of each Mining Ship relative to the Earth must be no larger than 6.0 km/s at the moment of launch and unloading resources. The velocity of each Mining Ship at the moment of rendezvous must be equal to the velocity of target asteroid.
- The time interval between two successive rendezvous of the same asteroid must be no less than 1 year: $t_{2i} - t_{1i} \ge 1 \text{ yr}$, where $t_{2i}$ and $t_{1i}$ denote the moments of the second and first rendezvous to the $i$-th asteroid, respectively. The collected mass $M_i$ on the $i$-th asteroid is constrained by $M_i \le k(t_{2i} - t_{1i})$ where the rate of mining resources is $k = 10.0 \text{ kg/yr}$.
- The initial mass of each Mining Ship is constrained by $m_0 = m_d + m_p + I m_s \le 3000 \text{ kg}$, where $m_d = 500 \text{ kg}$ is the constant dry mass, $m_p$ is the propellant mass, $I (\le 20)$ is the number of miners, and $m_s = 40 \text{ kg}$ is the mass of a miner. The final mass of each Mining Ship must be no smaller than the sum of the dry mass 500 kg and collected mass on the ship.
- The number of Mining Ships used in each solution, $N$, is constrained by
   $$ N \le \min \left[ 100, 2 \exp(\rho \bar{M}) \right], \text{ where } \rho = 0.004 \text{ kg}^{-1}, \text{ and } \bar{M} = \frac{1}{N} \sum_{i=1}^{60000} M_i \text{, in kilograms, is the} $$
   average collected mass transported back to the Earth per Mining Ship in this solution. Illustrative examples for the maximum number of Mining Ships are given in Table 1.

   **Table 1 Maximum number of Mining Ships as a function of the average collected mass**
   | $\bar{M}$ (kg) | 100 | 300 | 500 | 700 | 900 | 1000 |
   |----------------|-----|-----|-----|-----|-----|------|
   | $N_{max}$      | 2   | 6   | 14  | 32  | 73  | 100  |

- Gravity assists at the planets Venus, Earth, and Mars are allowed. The GA impulse model is given in Section 6.2. If the Mining Ship flies by the Earth with a hyperbolic excess velocity no larger than 6 km/s, all resources carried by this ship will be instantaneously unloaded on the Earth.
g. The distance from the Sun to each Mining Ship must be no less than 0.3 AU: $r(t) \ge 0.3 \text{ AU}$.
h. Tolerances in the position, velocity, and mass of each Mining Ship at all events are 1,000 km, 1.0 m/s, and 0.001 kg, respectively.

## 5. Appendices

### 5.1 Dynamic Model

In the Sun's central gravitational field, the motions of planets, asteroids, and un-propelled Mining Ships are all governed by
$$ \dot{\mathbf{r}} = \mathbf{v} $$
$$ \dot{\mathbf{v}} = -\frac{\mu}{r^3}\mathbf{r} $$
where $\mathbf{r}$ (with $r = ||\mathbf{r}||$) and $\mathbf{v}$ are the position and velocity in the J2000 heliocentric ecliptic frame, respectively, and $\mu$ is the gravitational parameter of the Sun.
The initial states of planets and asteroids are described by their initial mean anomaly $M_0$ along with their other five constant orbital elements: the semi-major axis $a$, eccentricity $e$, inclination $i$, longitude of ascending node $\Omega$, and argument of perihelion $\omega$. The mean anomaly at time $t$, $M_t$, is computed by
$$ M_t = M_0 + \sqrt{\frac{\mu}{a^3}}(t-t_0) $$
The orbital eccentric anomaly $E$ at time $t$ is obtained by solving the Kepler's equation:
$$ M_t = E - e \sin E $$
The orbital elements are related to the Cartesian position and velocity according to the expressions
$$ \mathbf{r} = r(\mathbf{P}\cos f + \mathbf{Q} \sin f) $$
$$ \mathbf{v} = \sqrt{\frac{\mu}{a(1-e^2)}} [-\mathbf{P}\sin f + \mathbf{Q}(e+\cos f)] $$
where $r = \frac{a(1-e^2)}{1+e\cos f}$, $\tan \frac{f}{2} = \sqrt{\frac{1+e}{1-e}} \tan \frac{E}{2}$, and $f$ is the true anomaly. The vectors $\mathbf{P}$ and $\mathbf{Q}$ are given by
$$ \mathbf{P} = \begin{pmatrix} \cos\omega\cos\Omega - \sin\omega\sin\Omega\cos i \\ \cos\omega\sin\Omega + \sin\omega\cos\Omega\cos i \\ \sin\omega\sin i \end{pmatrix} $$
$$ \mathbf{Q} = \begin{pmatrix} -\sin\omega\cos\Omega - \cos\omega\sin\Omega\cos i \\ -\sin\omega\sin\Omega + \cos\omega\cos\Omega\cos i \\ \cos\omega\sin i \end{pmatrix} $$
Every Mining Ship is propelled by the electric thrust, and its position, velocity, and mass change continuously when the thruster is on:
$$ \dot{\mathbf{r}} = \mathbf{v} $$
$$ \dot{\mathbf{v}} = -\frac{\mu}{r^3}\mathbf{r} + \frac{T}{m}\mathbf{\alpha} $$
$$ \dot{m} = -\frac{T}{I_{sp}g_0} $$
where $T$ ($0 \le T \le T_{max}$) is the magnitude of thrust, $\mathbf{\alpha}$ is the unit vector of thrust direction, $m$ is the instantaneous mass of the Mining Ship, $I_{sp}$ is the specific impulse, and $g_0$ is the gravitational acceleration at sea level. The Mining Ship mass changes instantaneously at the moment of rendezvous as it releases a miner or collects resources:
$$ m(t_{1i}^+) = m(t_{1i}^-) - m_s $$
$$ m(t_{2i}^+) = m(t_{2i}^-) + M_i $$
where $t_{1i}$ and $t_{2i}$ denote the moments of the first and second rendezvous with the $i$-th asteroid, respectively, $m_s$ is the mass of a miner, $M_i$ is the collected mass on the $i$-th asteroid, and $t^-$ and $t^+$ refer to the states immediately before and after the rendezvous, respectively.

### 5.2 GA impulse model

The GA impulse model (i.e., the same GA model used in the previous GTOCs) is used at the planetary flyby moment $t_g$ when the Mining Ship's positions at the moments immediately before and after GA, $\mathbf{r}(t_g^-)$ and $\mathbf{r}(t_g^+)$, respectively, are both equal to the planet's position $\mathbf{r}_p(t_g)$:
$$ \mathbf{r}(t_g^-) = \mathbf{r}(t_g^+) = \mathbf{r}_p(t_g) $$
The time spent inside the planetary influence is neglected. The hyperbolic excess velocity is calculated by the heliocentric velocity of the Mining Ship relative to that of the planet:
$$ \mathbf{v}_\infty(t_g^-) = \mathbf{v}(t_g^-) - \mathbf{v}_p(t_g) $$
$$ \mathbf{v}_\infty(t_g^+) = \mathbf{v}(t_g^+) - \mathbf{v}_p(t_g) $$
where $\mathbf{v}_p$ is the planet's velocity. The magnitude of hyperbolic excess velocity remains invariant, but its direction has an instantaneous change:
$$ ||\mathbf{v}_\infty(t_g^-)|| = ||\mathbf{v}_\infty(t_g^+)|| = v_\infty $$
$$ \mathbf{v}_\infty(t_g^-) \cdot \mathbf{v}_\infty(t_g^+) = v_\infty^2 \cos\theta $$
where the direction change angle $\theta$ is constrained by the minimum pericenter radius $r_{p,min}$
$$ \sin \frac{\theta}{2} = \frac{\mu_p/r_p}{v_\infty^2 + \mu_p/r_p}, \quad r_p \ge r_{p,min} $$
where $r_p$ is the pericenter radius and $\mu_p$ is the gravitational parameter of the planet.
The Mining Ship mass changes instantaneously at the moment of Earth flyby with a hyperbolic excess velocity no larger than 6.0 km/s:
$$ m(t_g^+) = m(t_g^-) - M $$
where $M$ is equal to the total collected mass carried by this ship.

### 5.3 Constants

The planets' parameters and orbital elements at the mission initial time $t = 64328$ MJD are given in Table 2. The values of some constant parameters are provided in Table 3.

**Table 2 The planets' parameters and orbital elements ($t = 64328$ MJD)**

|                                      | Venus                 | Earth                 | Mars                  |
|--------------------------------------|-----------------------|-----------------------|-----------------------|
| Gravitational parameter, km³/s²      | 3.24858592000e5       | 3.98600435436e5       | 4.28283752140e4       |
| Minimum pericenter radius, km        | 6351.0                | 6678.0                | 3689.0                |
| Semi-major axis, km                  | 1.08208010521e8       | 1.49579151285e8       | 2.27951663551e8       |
| Eccentricity                         | 6.72988099539e-3      | 1.65519129162e-2      | 9.33662184095e-2      |
| Inclination, deg                     | 3.39439096544         | 4.64389155500e-3      | 1.84693231241         |
| Longitude of ascending node, deg     | 7.65796397775e1       | 1.98956406477e2       | 4.94553142513e1       |
| Argument of perihelion, deg          | 5.51107191497e1       | 2.62960364700e2       | 2.86731029267e2       |
| Mean anomaly, deg                    | 1.11218416921e1       | 3.58039899470e2       | 2.38232037154e2       |

**Table 3 Values of some constant parameters**

| Parameter | Value             | Unit    |
|-----------|-------------------|---------|
| $\mu$     | 1.32712440018e11  | km³/s²  |
| $g_0$     | 9.80665           | m/s²    |
| AU        | 1.49597870691e8   | km      |
| Day       | 86400.0           | s       |
| Year      | 365.25            | day     |


## Submission format

The solution file contains all trajectories of Mining Ships. Each trajectory is expressed by its identification number (ship ID from 1 to $N$ ), event ID (from $^ { - 4 }$ to 60,000), epochs (t, MJD), and the histories of position (km), velocity $\left( \mathrm { k m } / \mathrm { s } \right)$ , thrust $( \mathrm { N } )$ , and mass $( \mathrm { k g } )$ . The vectors of position, velocity, and thrust should be projected onto the J2O00 heliocentric ecliptic Cartesian reference frame.The file format is summarized in Table 1 and described below. 

### Table 1 Solution File format

| Line num in file | Data |
|------------------|------|
| Line 1           | 1 &nbsp;&nbsp; 0 &nbsp;&nbsp; *t* &nbsp;&nbsp; rₓ &nbsp; rᵧ &nbsp; r_z &nbsp; vₓ &nbsp; vᵧ &nbsp; v_z &nbsp; m |
| Line 2           | 1 &nbsp;&nbsp; 0 &nbsp;&nbsp; *t* &nbsp;&nbsp; rₓ &nbsp; rᵧ &nbsp; r_z &nbsp; vₓ &nbsp; vᵧ &nbsp; v_z &nbsp; m |
| …                | … |
| Line *i*         | –1 &nbsp;&nbsp; *tᵢ* &nbsp;&nbsp; 0.0 &nbsp;&nbsp; 0.0 &nbsp;&nbsp; 0.0 |
| Line *i+1*       | –1 &nbsp;&nbsp; *tᵢ* &nbsp;&nbsp; Tₓ &nbsp;&nbsp; Tᵧ &nbsp;&nbsp; T_z |
| …                | … |
| Line *i+j+1*     | –1 &nbsp;&nbsp; *tᵢ+ⱼ* &nbsp;&nbsp; Tₓ &nbsp;&nbsp; Tᵧ &nbsp;&nbsp; T_z |
| Line *i+j+2*     | –1 &nbsp;&nbsp; *tᵢ+ⱼ* &nbsp;&nbsp; 0.0 &nbsp;&nbsp; 0.0 &nbsp;&nbsp; 0.0 |
| …                | … |
| Line *k*         | –2* &nbsp;&nbsp; *tₖ* &nbsp;&nbsp; rₓ &nbsp; rᵧ &nbsp; r_z &nbsp; vₓ &nbsp; vᵧ &nbsp; v_z &nbsp; m |
| Line *k+1*       | –2 &nbsp;&nbsp; *tₖ* &nbsp;&nbsp; rₓ &nbsp; rᵧ &nbsp; r_z &nbsp; vₓ &nbsp; vᵧ &nbsp; v_z &nbsp; m |
| …                | … |
| Line *p*         | 60000† &nbsp;&nbsp; *tₚ* &nbsp;&nbsp; rₓ &nbsp; rᵧ &nbsp; r_z &nbsp; vₓ &nbsp; vᵧ &nbsp; v_z &nbsp; m |
| Line *p+1*       | 60000 &nbsp;&nbsp; *tₚ* &nbsp;&nbsp; rₓ &nbsp; rᵧ &nbsp; r_z &nbsp; vₓ &nbsp; vᵧ &nbsp; v_z &nbsp; m |
| …                | … |
| Line *q*         | 2 &nbsp;&nbsp; 0 &nbsp;&nbsp; *t* &nbsp;&nbsp; rₓ &nbsp; rᵧ &nbsp; r_z &nbsp; vₓ &nbsp; vᵧ &nbsp; v_z &nbsp; m |
| Line *q+1*       | 2 &nbsp;&nbsp; 0 &nbsp;&nbsp; *t* &nbsp;&nbsp; rₓ &nbsp; rᵧ &nbsp; r_z &nbsp; vₓ &nbsp; vᵧ &nbsp; v_z &nbsp; m |
| …                | … |


The solution file must be an ASCI text file, and the format is defined in Data section of Table 1. Each line of the file has the following format: two integers followed by several floating point numbers.

The numbers must be separated by spaces.

The file can be divided into $N$ sections according to the ship ID (the first integer of each line increased from 1 to $N$ ). The first section and the beginning of the second section are shown in Table 1 as an example. Each section lists the events of launch from the Earth, burning arcs,rendezvous,and flybys, which are indicated by the event ID (the second integer of each line) in order of epochs (the first floating point number of each line). The second integer at the first line of each section is 0, indicating that this ship is leaving the Earth at its initial time.The second integer $^ { - 1 }$ represents burning arcs, $^ { - 2 }$ Venus flyby, $^ { - 3 }$ Earth flyby, $^ { - 4 }$ Mars flyby, and an asteroid ID represents rendezvous with this asteroid. Multiple lines with the same epochs are used to represents the events that occur at the same time. There are two types of formats for floating point numbers, depending on the events:

**a. Event $\mathbf { I D } = - \mathbf { 1 }$**

At the moment the thruster is switched either from shutdown to burning or vice versa, two lines must be listed with the same epochs but different thrusts. A burning arc is presented by starting with a line and ending with a line both with a zero thrust vector, and by evenly distributing lines with thrust components between the boundary lines. So, the second and penultimate lines have the same epochs with the first and last lines,respectively, but the thrust vectors are both nonzero. There is no need to provide data for coast arcs. The time interval between two successive lines in the burning arc has to be one day, and a partial-day increment from the antepenultimate line to the last two lines is permitted. Note that, when the solution is verified, the state and mass of each Mining Ship will be propagated by numerically integrating the differential equations of the state and the mass, of which the thrust will be approximated by a third Lagrange interpolating polynomial.

**b. Event $\mathbf { I D } = \mathbf { 0 } , - \mathbf { 2 } , - \mathbf { 3 } , - \mathbf { 4 }$ or asteroid ID**

The events of launches, rendezvous, and flybys are all represented by two lines consisting of the values of epochs, positions, velocities,and masses immediately before and after the events. The two lines corresponding to launch should be listed with the same epochs, positions, and masses,but with different velocities,of which the first one is the heliocentric velocity of the Earth,and the second one is the heliocentric velocity of the Mining Ship. The two lines corresponding to rendezvous should be listed with the same epochs, positions,and velocities, but with different masses as the ship unloads a miner or retrieves resources. The two lines corresponding to planetary flyby should be listed with the same epochs and positions, with the two heliocentric velocities of the Mining Ship satisfying the GA constraints,and with the mass difference for Earth flyby being equal to the mass of resources unloaded on the Earth. The errors in position, velocity, and mass will be evaluated by propagating the state and mass of the Mining Ship between two successive events (event $\mathrm { I D s } \ 0 , - 2 , - 3 , - 4$ or asteroid ID).


### First lines view of **GTOC12_Asteroids_Data.txt**

| ID | epoch (MJD) | a (AU)       | e           | i (deg)    | LAN (deg)   | argperi (deg) | M (deg)     |
|----|-------------|--------------|-------------|------------|-------------|---------------|-------------|
| 1  | 64328       | 3.073000e+00 | 1.177000e-01| 1.745000e+01 | 1.403000e+01 | 1.830000e+00  | 3.053207e+02 |
| 2  | 64328       | 3.193000e+00 | 2.341000e-01| 2.631000e+01 | 2.170900e+02 | 1.312800e+02  | 1.726297e+02 |
| 3  | 64328       | 3.142000e+00 | 5.460000e-02| 5.280000e+00 | 2.139500e+02 | 1.776000e+01  | 1.329012e+02 |

Format: The text file is a plain text table where columns are separated by spaces. The first row contains column headers, and subsequent rows contain numeric values, some in scientific notation. Each row corresponds to one record, with fields such as ID, epoch (MJD), semi-major axis (AU), eccentricity, inclination (deg), longitude of ascending node (deg), argument of periapsis (deg), and mean anomaly (deg).

# Memory



# Instructions

## Response format

Begin with a deep explanation of your solution in natural language, followed by.  
**one** markdown code block (```python … ```) containing the complete Python script.
The Python code must not use an if __name__ == "__main__": block; it should be written as a script that runs immediately when executed.
No extra headings or prose after the code block.

## Solution sketch guideline

- Focus on deep thinking: first decompose the problem into subproblems, then design the program architecture focusing only on the structure of the program and the functions. Remind that all the code must be in a unique python script.
- Required outputs (explicitly, in this order):
- 1. **Solution Sketch** — an enumerated decomposition of the problem into 8–15 atomic subproblems (each 3-5 sentences), and a 1–2 paragraph overall strategy explaining why the chosen pipeline balances speed and optimality. Think deeply about the plan since it is the key for success.
- 2. **Algorithmic Approach** — for each subproblem, propose 1–2 concrete algorithmic choices (cheap baseline + a higher-fidelity method). For each subproblem give 1 cheap baseline method and 1 higher-fidelity method (e.g., Lambert + Edelbaum baseline; Sims-Flanagan or direct collocation refinement). For each method list expected computational cost (big-O or rough runtime scaling), known failure modes, and suggested default hyperparameters.
- 3. **code structure** — Think deeply in natural language about the functions the program will need to solve the problem.
- 4. **Python code block** -- Here you have to implement the plan. First you should copy all the variables and formulas given that you will need. Then code the funcions properly documented, and finally the rest of code.
- Finally include test code for every step of the code, Keep in mind that the time available and computing capacity are very limited, so your strategy should seek to minimize the number of candidates and complex calculations.
- To win, you must devise a creative plan to maximize the total weighted mass by creating a valid solution file named Result.txt, this file must follow the submission format, pay attention to it since is quite challenging.
- Asteroids data can be found at /home/data/GTOC12_Asteroids_Data.txt
- About Result.txt format: Be careful to dont log, or to later eliminate empty lines in Result.txt, typically a last character \n at the end of the file. That will make validator crash. Always print the total number of lines in Result.txt.
- Unit Consistency: Be cautious with the units of measurement for each magnitude, as different libraries may use varying units. Establish and enforce common units for the entire simulation (time, mass, distance).
- A short task description follows.


## Implementation guideline

- <TOTAL_TIME_REMAINING: 9.0hrs 59.0mins 59.97789931297302secs>
- <TOTAL_STEPS_REMAINI NG: 50>
- Your script must be contained on a single markdown code block (wrapped in ```) which implements the solution:
- 1. **Implement the trajectory-design algorithm** inside one Python block, There should be no additional headings or text in your response. Just natural language text followed by a newline and then the markdown code block.
- 2. **MANDATORY:** write the submission file to `./submission/Result.txt`
- (fallback `./working/Result.txt`) exactly following *GTOC12 Submission Format*.
- 3. After writing, perform:
- requests.post("http://localhost:5000/validate", files={"file": open(path, "rb")})
- then print the server’s message.
- 4. Keep total runtime under the allotted time limit.
- All input data (asteroid list) resides in `/home/data`.
- Remember: **the grader looks for ./submission/Result.txt** – do not forget this file!


## Installed Packages

Your solution may leverage any of the installed packages such as: `plotly`, `skyfield`, `PyKEP`, `astropy`, `Dymos`, `scipy`, `JAX`, `PyGMO`, `SpiceyPy`, `matplotlib`, `TudatPy`, `pandas`, `CasADi`, `numpy`, `poliastro`. All listed libraries are pre-installed. No additional packages can be installed and there is **no internet**.

# Data Overview

```
best_solution/

best_submission/

input/
    GTOC12_Asteroids_Data.txt (60001 lines)
    de432s.bsp (10.9 MB)
    description.md (162 lines)
submission/

working/
\end{Verbatim}
\end{promptbox}

\section{Appendix: LLM-as-a-Judge Evaluation Prompt}
\label{appendix:judge_prompt}

This appendix contains the verbatim prompt used by the evaluator model (Gemini 2.5 Pro) to score the generated strategy drafts. The rubric is based on the 26 criteria established in collaboration with domain experts. 

\textit{Note: While the experimental design initially intended for iterative refinement of this prompt based on a comparison between LLM judgments and human expert evaluations, logistical constraints prevented the acquisition of these expert scores in time for the final iteration. Consequently, the scoring relies on the semantic interpretation of the expert-designed rubric by the judge model.}

\begin{promptbox}{Judge Prompt: Strategic Viability Assessment}
\begin{Verbatim}[
breaklines=true, 
    breakanywhere=true, 
    breaksymbolleft={}, 
    breakindent=1.5em, 
    fontsize=\small,     % Replacement for basicstyle
    fontfamily=tt        % Ensures monospaced typewriter font
]
You are an expert judge for the GTOC 12 Global Trajectory Optimization Competition.
Evaluate the provided Strategy Plan and Code Snippets based on the following 5 sections.
For each section, provide a brief justification and a True or False.

SECTION 1: Understanding of the objective function
- Understand that to maximize mined mass, miners need to be deployed ASAP.
- Understand that collection should happen as late as possible.
- Understand that the number of missions allowed depends on total mined mass.
- Understand that transfer time savings have differing effects on mined mass.

SECTION 2: Asteroid Accessibility and Filtering
- Filter inaccessible/costly asteroids (E->A, A->E).
- Robust methods to rank asteroids based on transfer cost.
- Group asteroids reachable early in the mission.
- Group asteroids that can return late (optimizing phasing).
- Strategies to cluster asteroids for ease of transfer (A->A).

SECTION 3: Low-Thrust Transfer Estimation
- Recognition that high-fidelity trajectories are too slow for preliminary design.
- Propose fast, approximate methods for feasibility checks without full optimization.
- Method to optimize low-thrust transfer with required accuracy.
- Method of transcribing low-thrust transfer to solution file format.

SECTION 4: Mission Architecture and Logical Constraints
- Minimum feasible structure (Earth–Asteroid–wait–Earth).
- Simplification via single mothership for deploy/collect.
- Constraint: spacecraft count depends on mined mass.
- Phasing: Deploy early, collect late.
- Establish reasonable TOF limits (E-A, A-E, A-A).
- Define realistic number of asteroids per mothership.
- Algorithm to build feasible transfer sequences (combinatorial solver).
- Scalable method to transform coarse estimates to high-fidelity.

SECTION 5: Coupling and Optimization of Multi-Mission Systems
- Account for interdependencies (deployers/collectors not self-sufficient).
- Determine relevance of Multi-Gravity Assist (MGA).
- Fully optimize independent missions.
- Jointly optimize interdependent missions.
- Algorithm to select from fully optimized motherships to maximize objective function.
\end{Verbatim}
\end{promptbox}

\end{document}